\documentclass[conference]{IEEEtran}
\IEEEoverridecommandlockouts
\usepackage{cite}
\usepackage{amsmath,amssymb,amsfonts}
\usepackage{algorithmic}
\usepackage{graphicx}
\usepackage{textcomp}
\usepackage{xcolor}
\def\BibTeX{{\rm B\kern-.05em{\sc i\kern-.025em b}\kern-.08em
    T\kern-.1667em\lower.7ex\hbox{E}\kern-.125emX}}
\begin{document}

\title{Convolutional Neural Network and Adversarial Autoencoder in EEG images classification\\
}
\author{
		\IEEEauthorblockN{Albert Nasybullin}
		\IEEEauthorblockA{\textit{Faculty of Computer Science}\\
		\textit{and Engineering,}\\
			\textit{Innopolis University}\\
			Innopolis,	Russia\\
			a.nasibullin@innopolis.university}
		\and
        \IEEEauthorblockN{Semen Kurkin}
		\IEEEauthorblockA{\textit{Neuroscience and Cognitive}\\ 
			\textit{Technology Laboratory,}\\
			\textit{Center for Technologies in Robotics}\\
			\textit{and Mechatronics Components,}\\
			\textit{Innopolis University}\\
			Innopolis,	Russia\\
			kurkinsa@gmail.com}
		}

\maketitle

\begin{abstract}
In this paper, we consider applying computer vision algorithms for the classification problem one faces in neuroscience during EEG data analysis. Our approach is to apply a combination of computer vision and neural network methods to solve human brain activity classification problems during hand movement. We pre-processed raw EEG signals and generated 2D EEG topograms. Later, we developed supervised and semi-supervised neural networks to classify different motor cortex activities.
\end{abstract}


\section{Introduction}
Brain signals classification is one of the essential problems in neuroscience \cite{chholak2019phase, kurkin2019machine, Gordleeva2020, kurkin2021oxygen, bukina2021modern, chholak2021event, maksimenko2021effect, kurkin2020system, kurkin2019artificial}. The typical approach for analysis and classification of human brain motor activity collected by EEG are Linear Discriminant Analysis (LDA), Support Vector Machine (SVM), and K-Nearest Neighbors (k-NN) classifier \cite{cha2019analysis, Gordleeva2017, Liburkina2018, Grigorev2020, badarin2020studying}. Neural network-based classification is also widely used. It includes Artificial Neural Networks (ANN), Convolutional Neural Networks (CNN), Deep Neural Networks (DNN), and a lot more \cite{andreev2019synchronization, andreev2021synchronization, ponomarenko2021assessment}.

In this work, we work with data collected by EEG, using wavelet analysis techniques to generate a dataset of topograms, and implement neural network-based solutions for brain motor activity classification problems.

\section{Data collection and data structure}

Participants of the experiment were fifteen volunteers. All healthy males at the age from 20 to 45 years. Participants were right-handed, non-smokers, and regularly engage in physical activity. None of the participants has been diagnosed with musculoskeletal or central nervous system diseases or prescribed medication. For the sake of the study, the participants followed a healthy lifestyle for at least 48-hours before the experiment. That includes 8-hours long night rest, abstinence from alcohol, a reasonably limited caffeine consuming and physical exercises. 

Before the experiment has begun, all participants were notified about the goals, methods, and possible discomforts of the experiment. The procedure followed Helsinki's Declaration and was approved by the Ethics Committee of Innopolis University.

\subsection{Task}

The procedure of the experiment is next. The subject comfortably sits in a chair, his hands are on armrests. At the beginning and the end of each participant's session 3 minutes long background brain activity was recorded. During background recording participants were told try to relax and don not move their hands. Active phase of the experiment includes right or left hand movements, instructions to start movements were provided on a screen. We collected 20 trials of movements for each hand (40 trials in total).

Every individual trial includes number of phases accompanied with commands on the screen. First, trial starts with attention fixation. There is a signal for participant to prepare for an experiment begins, a bright cross lights up on a black screen for 2 seconds. Second, the cross stays on the screen, left- or right-oriented arrow appears on the top of it for 1.5 seconds. Third, the arrow disappears from the screen and the participant performs left hand movement or right hand movement correspondingly. In the provided experiment the hand movement is bending and unbending of fingers to the center of palm. At the end, there is a rest phase: the cross disappears, a participant is seeing a black screen for 15 seconds and resting before a next command shows up.

\subsection{EEG data acquisition and preprocessing}
EEG signals were recorded using the actiCHamp electroencephalograph manufactured by Brain Products, Germany. EEG signals were recorded with 32 channels in accordance with the 10-10 scheme. The ground was located at the site of the Fpz electrode, and the reference electrode was placed behind the right ear. For EEG registration, active Ag/AgCl electrodes ActiCAP were used, which were located on the scalp surface in the sockets of a special EasyCAP cap. To improve signal quality and provide better conductivity, the scalp was pretreated with NuPrep abrasive gel, and then the electrodes were positioned using SuperVisc conductive gel. During the experiment, the conductivity values were monitored at each of the EEG electrodes. Typically, the values were $<$ 25~k$\Omega$  which is sufficient for the correct operation of active EEG electrodes. The raw EEG signals were sampled at 1000 Hz and filtered by a 50--Hz notch filter by an embedded hardware-software data acquisition complex. Additionally, raw EEG signals were filtered by the 5th-order Butterworth filter with cut-off points at 1 Hz and 100 Hz. Eyes blinking and heartbeat artifacts removal was performed by the Independent Component Analysis (ICA). Data was then inspected manually and corrected for remaining artifacts.

\subsection{Data processing}\label{AA}
After the experiment, we have raw EEG data for 15 human subjects and 20 trials for each human subject. 

The next step is to convert raw EEG data into 2D human scalp topographies and to choose correct samples from data trials. There is no formally verified correct way to do that. In this work, we empirically discovered the next sample selection strategy, which led to satisfying results:

\begin{itemize}
    \item Chosen frequency is "mu" (9-11 Hz). "Mu" frequency was chosen because of the nature of the experiment. It is well known, that "mu"/"alpha" frequencies are responsible for motor activity.
    \item To avoid edge effects, we cut the following intervals out:
            \begin{itemize}
                \item 5.0-5.5 seconds,
                \item 8.5-10.0 seconds.
            \end{itemize}
    \item Time frame 0.0-5.0 seconds was not used. This interval corresponds to the recording of ``baseline'' activity.
    \item To extract the maximum amount of useful data we use ``slicing windows'' as a strategy to export EEG topographies:
            \begin{itemize}
                \item 5.5-7.0 seconds,
                \item 6.0-7.5 seconds,
                \item 6.5-8.0 seconds,
                \item 7.0-8.5 seconds.
            \end{itemize}
    \item To double the exported amount of data we use two different baseline procedures:
            \begin{itemize}
                \item ``absolute'',
                \item ``relative''.
            \end{itemize}
\end{itemize}

To generate EEG topographies Matlab's package FieldTrip Toolbox was used \cite{maris2007nonparametric}.

\subsection{Finalized dataset}

The final dataset consists of 939 images and two classes. The resolution of images is 840 by 630 pixels. Dataset structure is represented in Figs.~1 and 2. Approximately data was split in 80/20 proportion. 80$\%$ of images are train set, 20$\%$ of images are test set. The current data split was chosen according to Scaling Law \cite{guyon1997scaling} splits.

\begin{figure}[htbp]
\centerline{\includegraphics[width=0.3\textwidth]{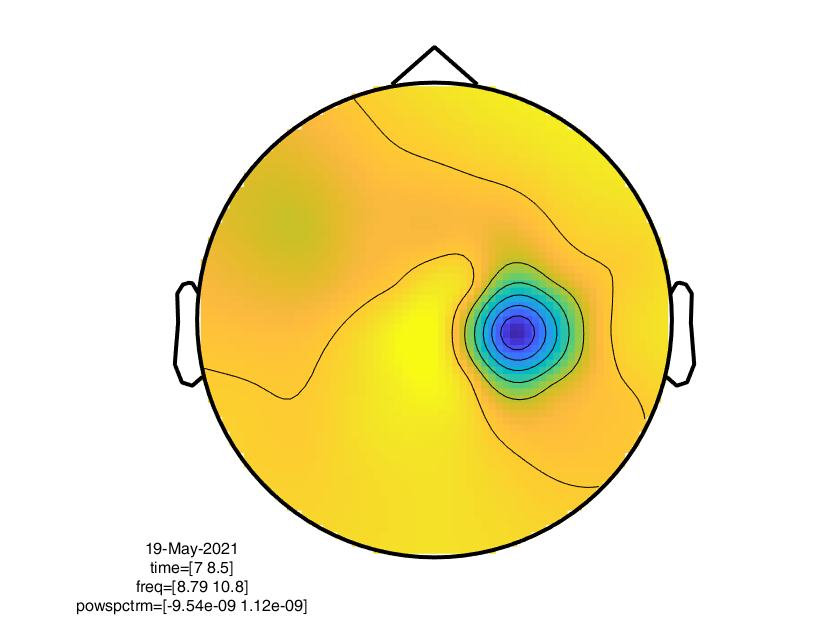}}
\caption{Left hand-related EEG topography in Mu frequency band.}
\label{left10}
\end{figure}

\begin{figure}[htbp]
\centerline{\includegraphics[width=0.3\textwidth]{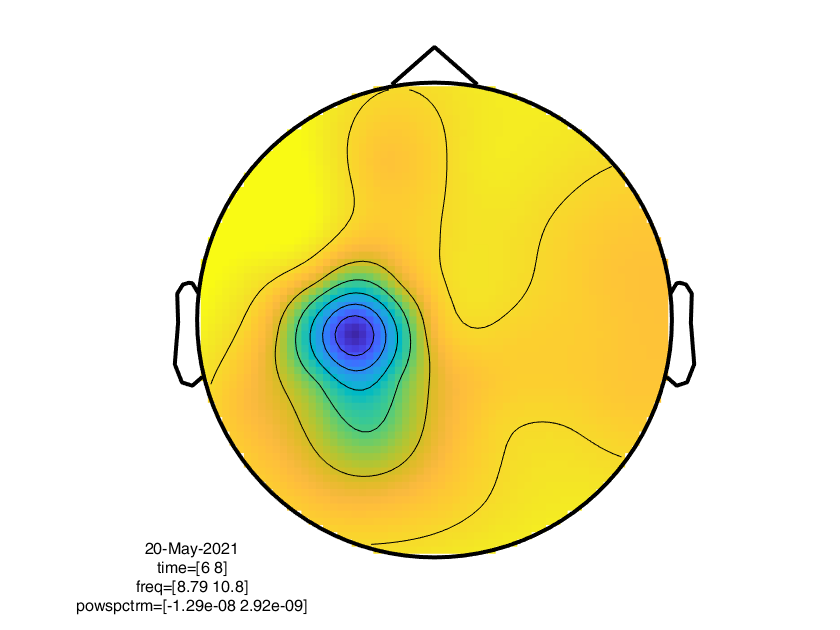}}
\caption{Right hand-related EEG topography in Mu frequency band.}
\label{right200}
\end{figure}

\section{Convolutional Neural Network (CNN)}

Input shape for Convolutional Neural Network is 84 by 63 pixels. Reshaped images tended to better quality performance than original images with the same network structure \cite{valueva2020application}. Other parameters of the network were configured empirically during the development. 

The current network uses a 5 by 5 kernel. The network contains four convolutional layers with increasing density, four pooling layers with dimensions 2 by 2, three fully connected layers with decreasing density to the number of classes we were looking for.

ReLU (Rectified Linear Unit) was chosen as an activation function. ReLU was selected for the sake of reducing computational expenses and compensating high dimensionality of other parts of the designed network. To the last fully connected layer Softmax activation function was used. 
ReLU activation function is defined as the positive part of its argument. Here, $x$ is a input of neuron:

\begin{equation} 
Relu(x) = max(0,x)
\end{equation}

SoftMax is commonly used as an activation function for the last layer of Artificial Neural Networks (ANN). SoftMax uses Luce's choice axiom and normalizes the output of the network to a probability distribution over predicted output classes. Input of SoftMax is a vector $x$ of $K$ amount of real numbers:

\begin{equation}
\text{Softmax}(x_{i}) = \frac{\exp(x_i)}{\sum_{j}^{K} \exp(x_j)}
\end{equation}

\subsection{Results}

Convolutional Neural Network performance is satisfying for our classification problem. Supervised learning on just 10 epochs reaches an accuracy of 93,75$\%$. Algorithm performance on a larger number of epochs is unpredictable but a bigger size train dataset may lead to better performance.

\begin{figure}[htbp]
\centerline{\includegraphics[width=0.4\textwidth]{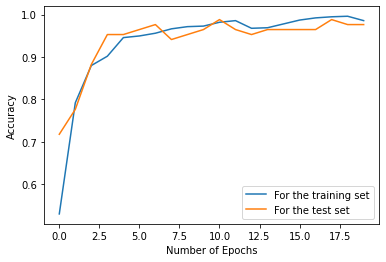}}
\caption{The CNN's accuracy is satisfying. The network reaches accuracy over 90$\%$even on small-scale data.}
\label{CNNacc}
\end{figure}

\section{Semi-supervised Adversarial Autoencoder (AAE)}

The main idea of GAN-based methods is a competition between two objects: a generator $G$ and a discriminator $D$. Generator is trying to create images that look like they belong to the original dataset $X$. The work of the discriminator is to distinguish between original data $X$ and generated images $G(X)$. In the best-case scenario, the training stops when the generator can outplay (``fool'') the discriminator. So, the generalized training objective for GAN can be described as:

\begin{equation}
\begin{split}
\min _{G} \max _{D} V(D, G)=\mathbb{E}_{x \sim p_{\text {data }}(x)}[\log D(x)]- \\
-\mathbb{E}_{z \sim p_{z}(z)}[\log D(G(z))]
\end{split}
\end{equation}

In the current architecture, we use two sub-networks made up for autoencoders \cite{vu2019anomaly}. The training objective for the discriminator sub-network is to maximize a pixel-wise error between the reconstructed image from original dataset $X$ and generated image $G(X)$:

\begin{equation}
\mathcal{L}_{D}=\|X-D(X)\|_{1}-\|G(X)-D(G(X))\|_{1}
\end{equation}

At the same time, the generator is trying to minimize the same error and ``fool'' the discriminator. Training objective for the generator may be represented as:

\begin{equation}
\mathcal{L}_{G}=\|X-D(X)\|_{1}+\|G(X)-D(G(X))\|_{1}
\end{equation}

We chose $L_g$ (pixel-wise error) to achieve sharper results, following insights from Isola \cite{isola2017image}. For the sake of having a more robust model, generator sub-network and discriminator sub-network are created only from fully convolutional layers.

\subsection{Results}

The network was trained at 400 epochs. The training results are satisfying for our goals. Still, the results of each training are unstable. Training results are floating. The network success rate floats from 60$\%$ to 68$\%$ during different training sessions. The best-achieved result is 68$\%$.

\begin{figure}[htbp]
\centerline{\includegraphics[width=0.4\textwidth]{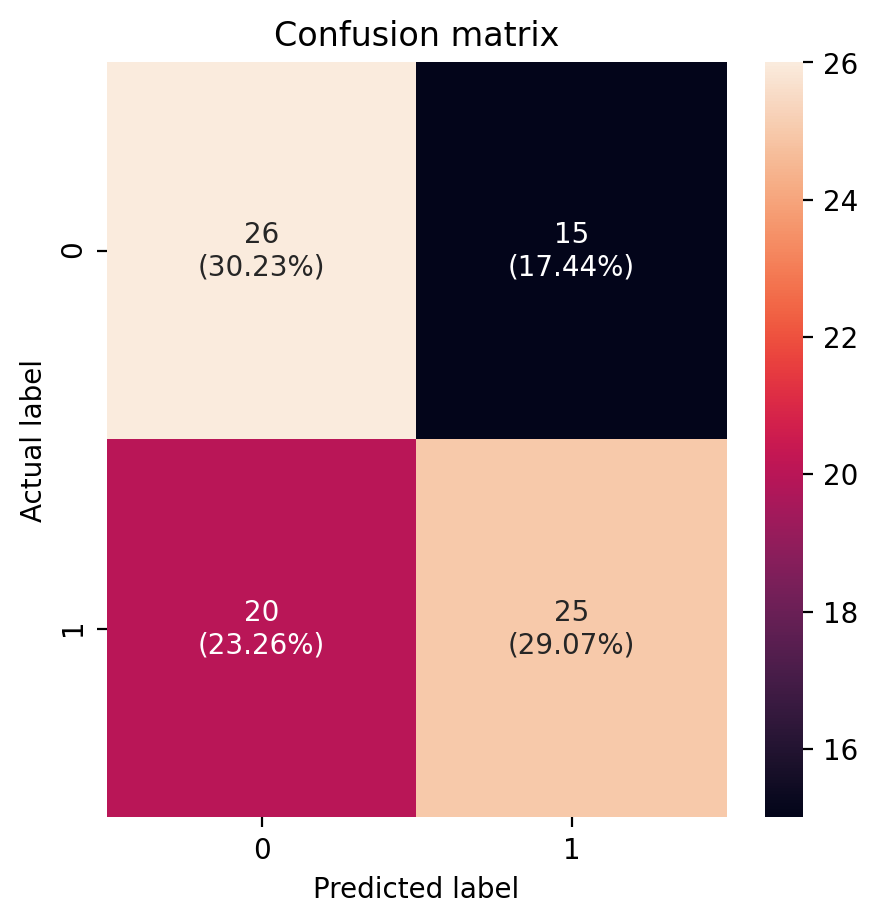}}
\caption{The Confusion Matrix of Adversarial Autoencoder. 1 = class "left hand", 0 = class "right hand".}
\label{confusionMatrix}
\end{figure}

\begin{figure}[!htbp]
\centerline{\includegraphics[width=0.4\textwidth]{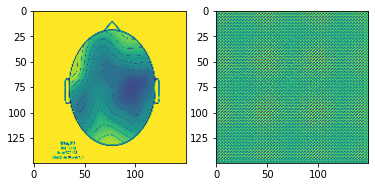}}
\caption{Original Image after the augmentation is on a left. Generated Image is on a right. Step 0.}
\label{generatorStep0}
\end{figure}

\begin{figure}[!htbp]
\centerline{\includegraphics[width=0.4\textwidth]{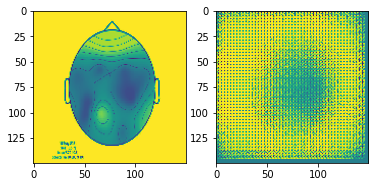}}
\caption{Original Image after the augmentation is on a left. Generated Image is on a right. Step 150.}
\label{generatorStep150}
\end{figure}

\section{Conclusion}
In the work, we explored opportunities and capabilities of neural networks and computer vision-based techniques in the analysis and classification of human brain motor cortex activity using EEG neuroimaging. We have tested both supervised and semi-supervised approaches. As a result, motor cortex activity was successfully classified. However, the adversarial autoencoder approach requires more time and effort to achieve better results.


\end{document}